\documentclass[conference]{IEEEtran}
\IEEEoverridecommandlockouts
\usepackage{cite}
\usepackage{amsmath,amssymb,amsfonts}
\usepackage{algorithmic}
\usepackage{graphicx}
\usepackage{textcomp}
\usepackage{xcolor}
\def\BibTeX{{\rm B\kern-.05em{\sc i\kern-.025em b}\kern-.08em
    T\kern-.1667em\lower.7ex\hbox{E}\kern-.125emX}}
\begin{document}

\title{Classification of PS and ABS Black Plastics for WEEE Recycling Applications
\thanks{This work has been carried out by Anton Persson and Niklas Dymne in the context of their Bachelor Thesis at Halmstad University (Intelligent Systems program), with the support of STENA Nordic Recycling Center (SNRC) in Halmstad.}
}

\author{\IEEEauthorblockN{Anton Persson, Niklas Dymne, Fernando Alonso-Fernandez}
\IEEEauthorblockA{\textit{School of Information Technology (ITE)} \\
\textit{Halmstad University}, Sweden \\
antper18@student.hh.se, nikdym18@student.hh.se, feralo@hh.se}
}

\maketitle

\begin{abstract}
Pollution and climate change are some of the biggest challenges that humanity is facing. In such a context, efficient recycling is a crucial tool for a sustainable future. 
This work is aimed at creating a system that can classify different types of
plastics by using picture analysis, in particular, black plastics of the type Polystyrene (PS) and Acrylonitrile Butadiene Styrene (ABS).
They are two common plastics from Waste from Electrical and Electronic Equipment (WEEE).
For this purpose, a Convolutional Neural Network has been tested and retrained, obtaining a validation accuracy of 95\%.
Using a separate test set, average accuracy goes down to 86.6\%, but a further look at the results shows that the ABS type is correctly classified 100\% of the time, so it is the PS type that accumulates all the errors. 
Overall, this demonstrates the feasibility of classifying black plastics using CNN machine learning techniques. 
It is believed that if a more diverse and extensive image dataset becomes
available, a system with higher reliability that generalizes well could
be developed using the proposed methodology.
\end{abstract}


\section{Introduction}

Humanity is facing a big challenge in fighting against climate change and pollution.
A big part of counteracting its effect is expanding recycling facilities and
improving existing recycling processes.
A growing concern due to its increasing production and consumption is plastic recycling, since the majority of plastic is either landfilled or incinerated \cite{9319989}.
The effect is even more dramatic with Single-Use Plastic products (SUPs), which are used once, or for a short period of time, before being thrown away. 
Another source of plastic waste that is also increasing rapidly is Waste from Electrical and Electronic Equipment (WEEE), which comes from all kind of industrial and consumer equipment at the end of their life. 
WEEE waste contains a complex mixture of materials, some of which are hazardous, with legislations requiring a steep reduction of WEEE plastics going to landfill \cite{Buekens14}.

This work is a collaboration of Halmstad University with STENA Nordic Recycling Center (SNRC) in Halmstad. At SNRC, they recycle plastic, metal, rubber, and much more from electronics. 
In particular, they are interested in distinguishing different types of black plastics from electronics, which constitute the majority of plastics that they recycle. 
Currently, the task is done with mechanical dividers and a electrostate which applies an electric charge. Since each plastic reacts differently to the electric charge, it is projected to a different divider, making possible to separate it.  
However, this is sensitive to the flow (amount) of plastics that is being processed, the size of the plastics, or the moisture of the ambient. 
There is no method in place that can adapt in real time to variations in these factors, so if the electric charge or the position of the dividers is not correct, the plastics will end up in the wrong place at the end of the separation line.

The goal of this work is, by image analysis, classify two types of black plastics. The system shall first segment and then classify the plastics appearing in the image.
The two types are Polystyrene (PS) and Acrylonitrile Butadiene Styrene (ABS), which are the most common types of black plastic handled at SNRC. 
A Convolutional Neural Network (CNN) based on AlexNet has been retrained and evaluated for this purpose, given the huge success shown in recent years by CNNs in several object recognition and classification tasks \cite{[Lecun15],[Schmidhuber15]}. 
A dataset with 215 pictures of each type of plastic (PS and ABS) has been captured as well, which is used for training and evaluation.
Due to time limitation, the work of this paper will be limited to the fundamentals of the classification system, without physically implementing our concept at SNRC. 
We employ pictures of stationary plastics captured with smartphone camera and a ring light illuminating the scene. 
With this simple and cheap setup, we show the feasibility of the proposed solution in classifying black plastics.

\begin{figure*}[htb]
\centering
        \includegraphics[width=0.95\textwidth]{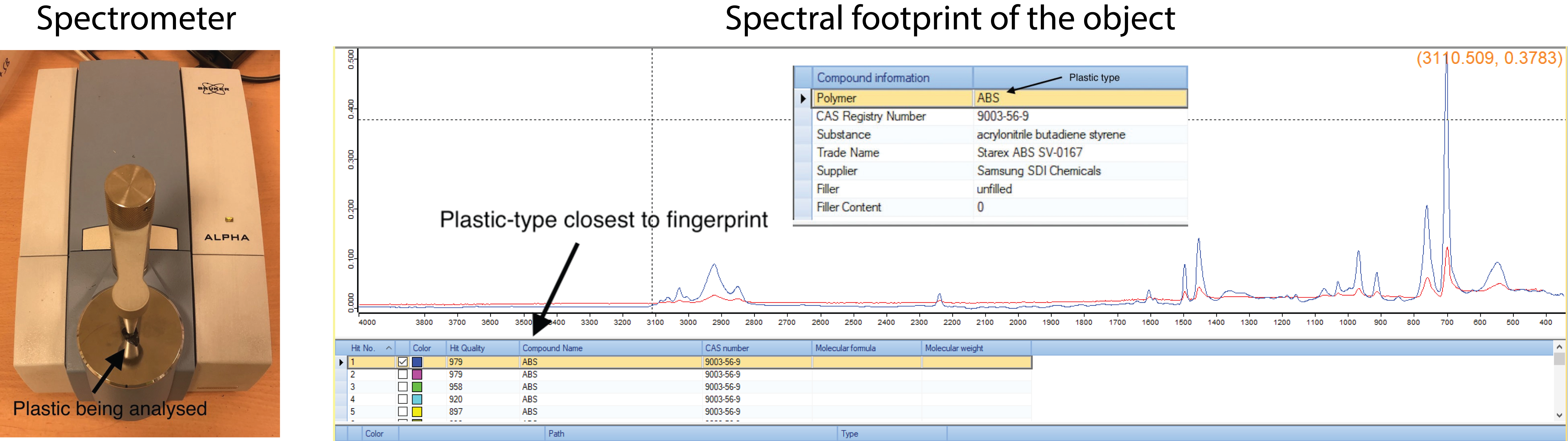}
        \caption{Black plastic separation using MWIR radiation.}
\label{fig3-6-7}
\end{figure*}

\section{Related Works}

\subsection{Using Near Infra-Red (NIR) illumination}

NIR sends beams in the wavelength of 1–1.7 $\mu$m, and by analysing the scattering from the plastic, it allows to specify its type \cite{1}. 
NIR is used to sort other types of materials as well, such as wood and metal, and it has other uses such as analysing if an egg is fresh without cracking it \cite{2}. 
NIR is also effective in detecting different types of WEEE plastics, as long as the infra-red beams scatter off the objects. 
This happens if the object is light in colours, such as white, yellow and other non-dark colours.
A problem occurs when the objects are dark. Since NIR light scatters poorly
from dark objects, it does not allow to properly classify them \cite{3}.
This poses a limitation in this paper, since our plastics of interest are
black.

\subsection{Using Mid Wavelength Infra-Red (MWIR) illumination}

MWIR operates at 3-12 $\mu$m, and it can determine different types of black plastics. %
A limitation is that it cannot operate on fast-moving objects, such as the plastics in the spread of electrostats \cite{1}.
The plastic piece has to be pressed against the spectrometer’s lens to determine
the plastic-type. 
This makes unfeasible to analyse a large amount of moving objects in a factory environment.
However, MWIR radiation will be used in the gathering of data for this paper, due to equipment available at SNRC (Figure~\ref{fig3-6-7}).
In addition, MWIR allows to separate moving objects from the background \cite{4}, providing a useful tool when our system is extended to an operating working environment.

\subsection{Using images in the visible (VW) range}

By analyzing shape or color of objects captured with regular (visible) cameras, it is possible to classify its type. 
For example, to sort plastic bottles by color \cite{11}.
Color analysis is also used in the food industry to analyze for example 
quality of meat \cite{14}, 
hardness of dates \cite{15}, 
potato chips category \cite{16},
or maturity of oranges \cite{17}. 
Although this does not need expensive equipment (just a regular camera and a sufficiently illuminated scene), 
we have faced issues when trying to classify black plastics, since its color is similar. 
Shape is not an alternative either, since the size or shape of plastics after dismantling electronic equipment is not consistent. 

Additionally, staff at SNRC said that they could determine the type of plastic by just looking at them, since its appearance and roughness looks different.
This sparked the idea that analyzing texture features instead of the color or the shape would be a better alternative. 
More precisely, we have explored data-driven learning of features, as enabled by Convolutional Neural Networks (CNNs).
CNNs are a powerful tool for computer vision, and they are used in the wast majority of image analysis applications in modern-day research \cite{[Lecun15],[Schmidhuber15]}.
To counteract over-fitting due to a reduced amount of training data, we explore transfer learning from a generic task from which a large amount of data is available \cite{[Russakovsky15_ImagenetChallege]}.

\subsection{Using Chemical Analysis}

To determine the type of plastic using chemical analysis, they can be spread out
on a piece of paper, and then different chemicals are poured on to the plastics.
Depending on the chemical, some plastic types will stick to the paper, while the rest will fall of if the paper is turned upside-down.
Although this method would allow very high accuracy, it is inefficient for our purposes.
Another possibility is the flotation technique \cite{6}. 
If the plastics have different densities, when they are immersed in an appropriate liquid, the less dense materials will float, and the more dense ones will sink.
It the plastics to separate have similar densities, bubbled liquid can be used instead in the so-called gamma floating technique. 
Based on the hydrophobicity (attraction to water) of each plastic type, the bubbles will stick to some types and not to others, allowing separation. 
This method is also available at SNRC, but its inefficiency in separating large amounts of plastic quickly also makes it impractical for our purposes.
In addition, it also alters the structure of the surface of the plastic, perturbing further analysis by image processing methods.

\subsection{Using Electrostatic Properties}

\begin{figure}[htb]
\centering
        \includegraphics[width=0.45\textwidth]{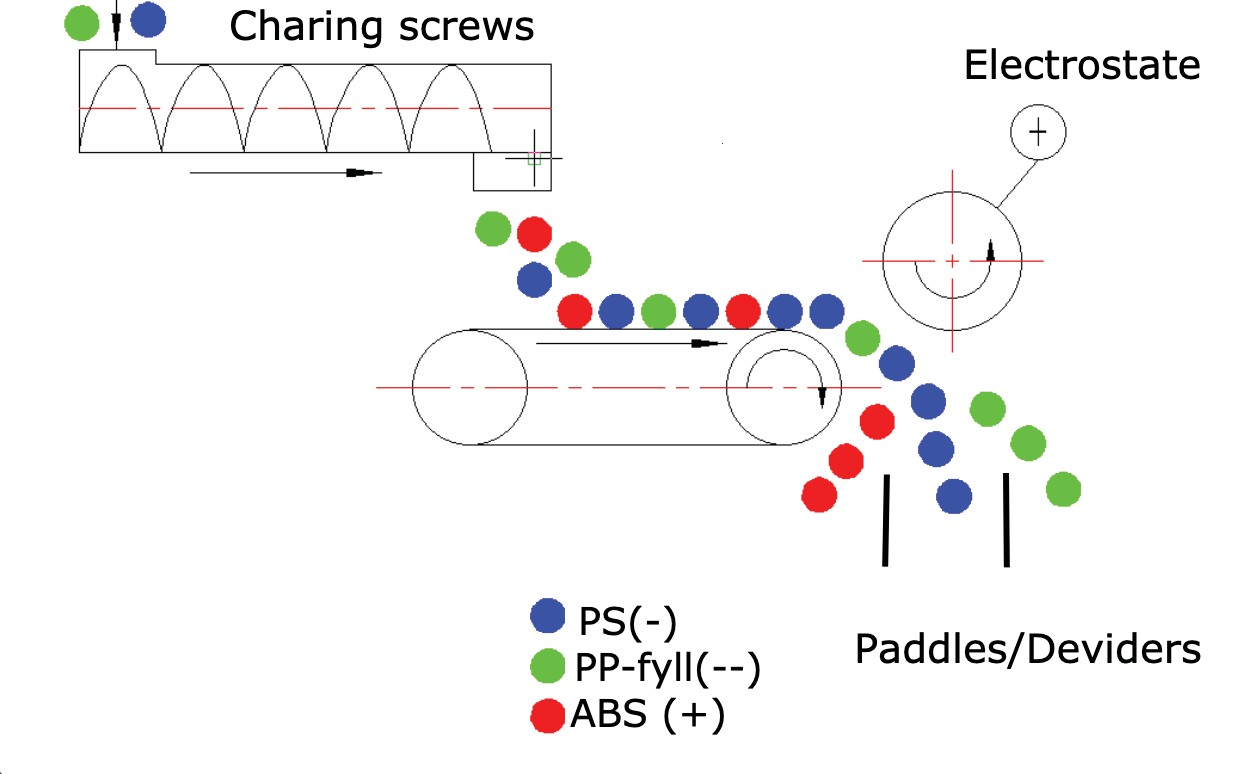}
        \caption{Black plastic separation using electrostatic properties.}
\label{fig1}
\end{figure}

This method is based on the different reaction of the plastics when an electric charge is applied to them (Figure~\ref{fig1}).  
At SNRC, they handle three types of black plastics: Acrylonitrile-butadiene-styrene (ABS), Polystyrene (PS), and Polypropene (PP-fyll).
However, we will concentrate on ABS and PS since they are the most common ones. 
When a positive charge is applied with an electrostate, ABS becomes positively charged (+), PS becomes negatively charged (-) and PP-fyll becomes the most negative of the three(–-). 
As a result, PS and PP-fyll are attracted to the electrostate, but with a different
force due to its different negative charge, while ABS is repelled. 
This method is available at SNRC but, as mentioned above, it is highly sensitive to the flow of plastics, its size, and other environmental factors such as moisture.

\begin{figure}[htb]
\centering
        \includegraphics[width=0.45\textwidth]{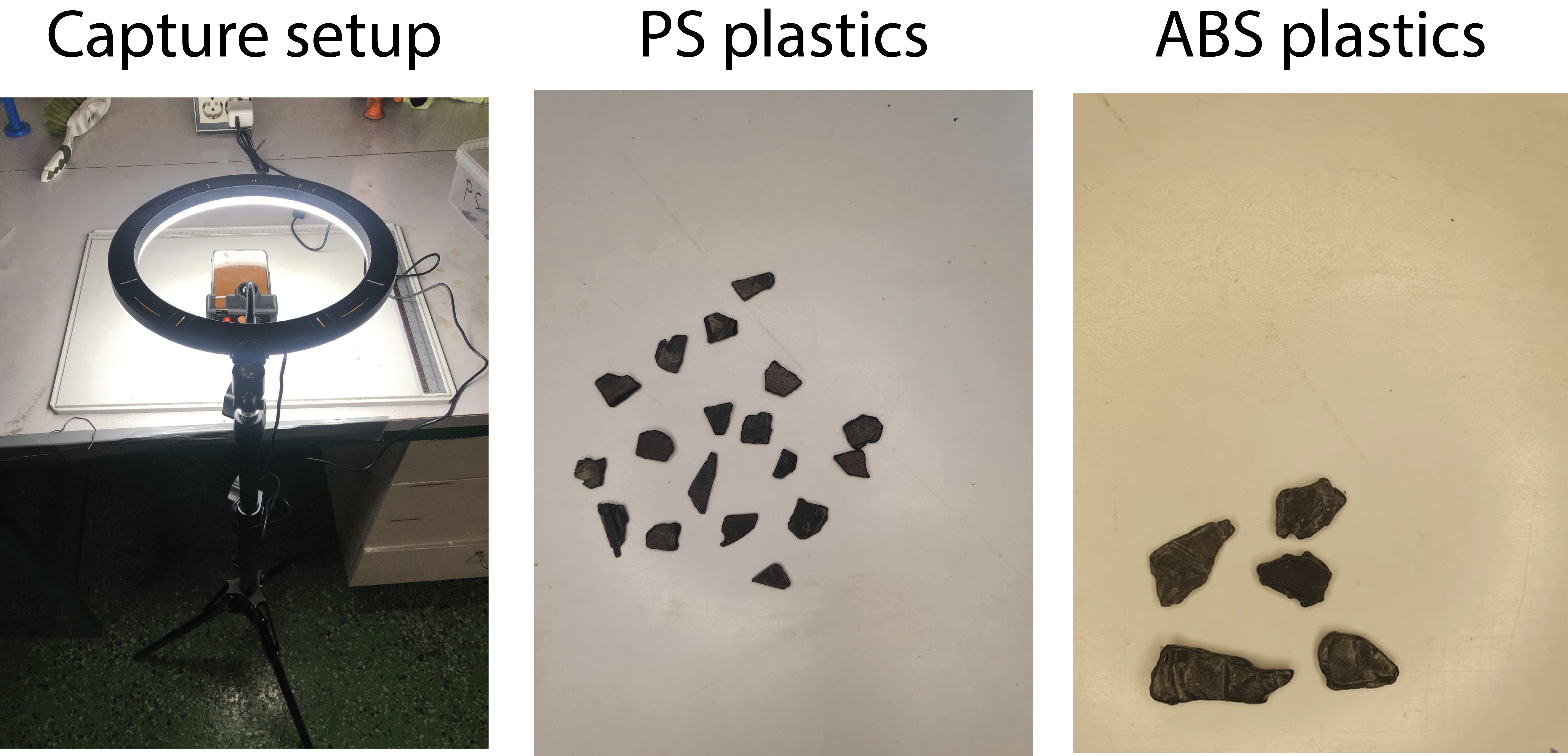}
        \caption{Data acquisition setup and example of captured images.}
\label{fig8-15-16}
\end{figure}

\begin{figure}[htb]
\centering
        \includegraphics[width=0.45\textwidth]{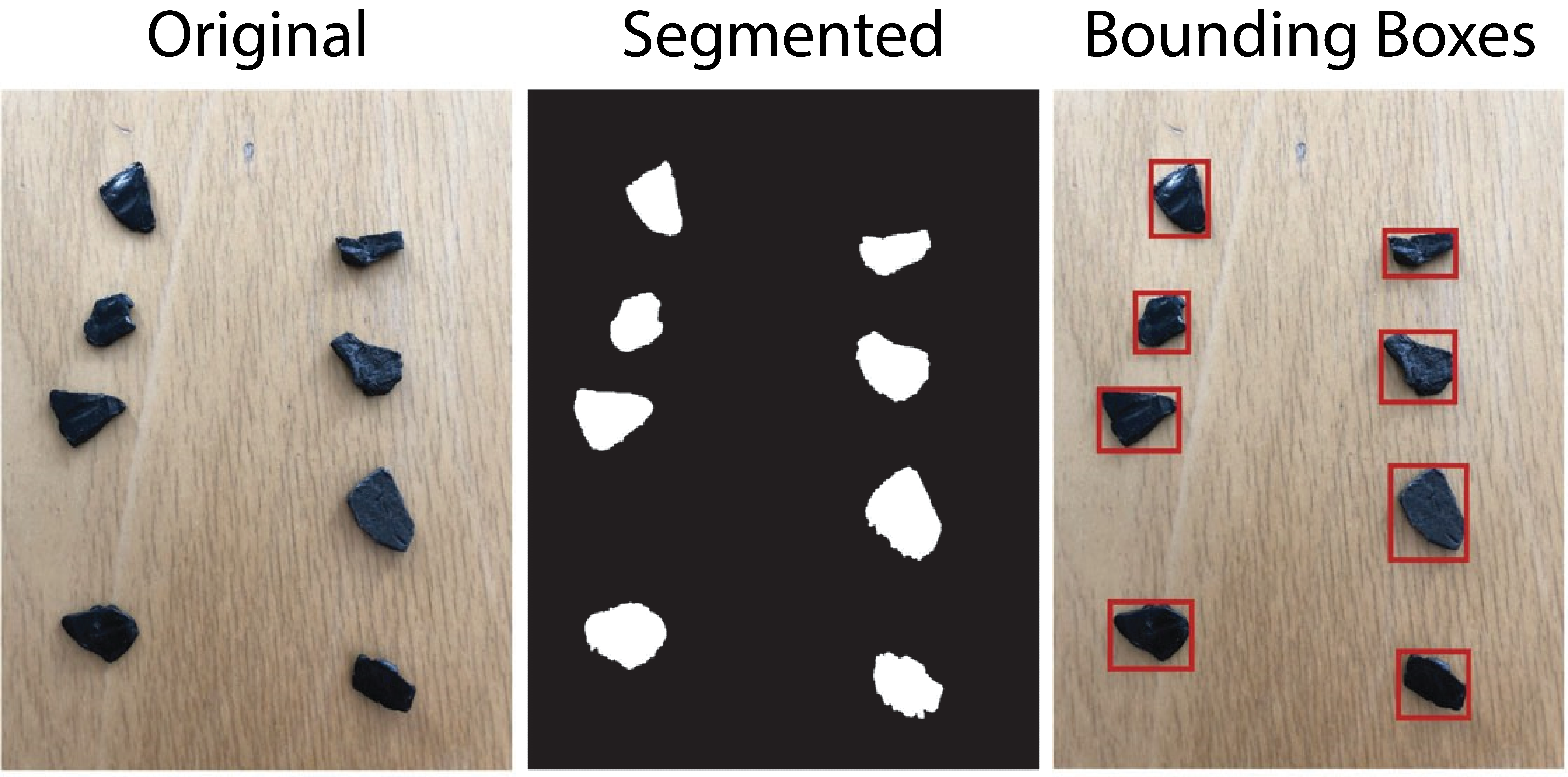}
        \caption{Example of plastic segmentation.}
\label{fig4}
\end{figure}

\begin{figure}[htb]
\centering
        \includegraphics[width=0.45\textwidth]{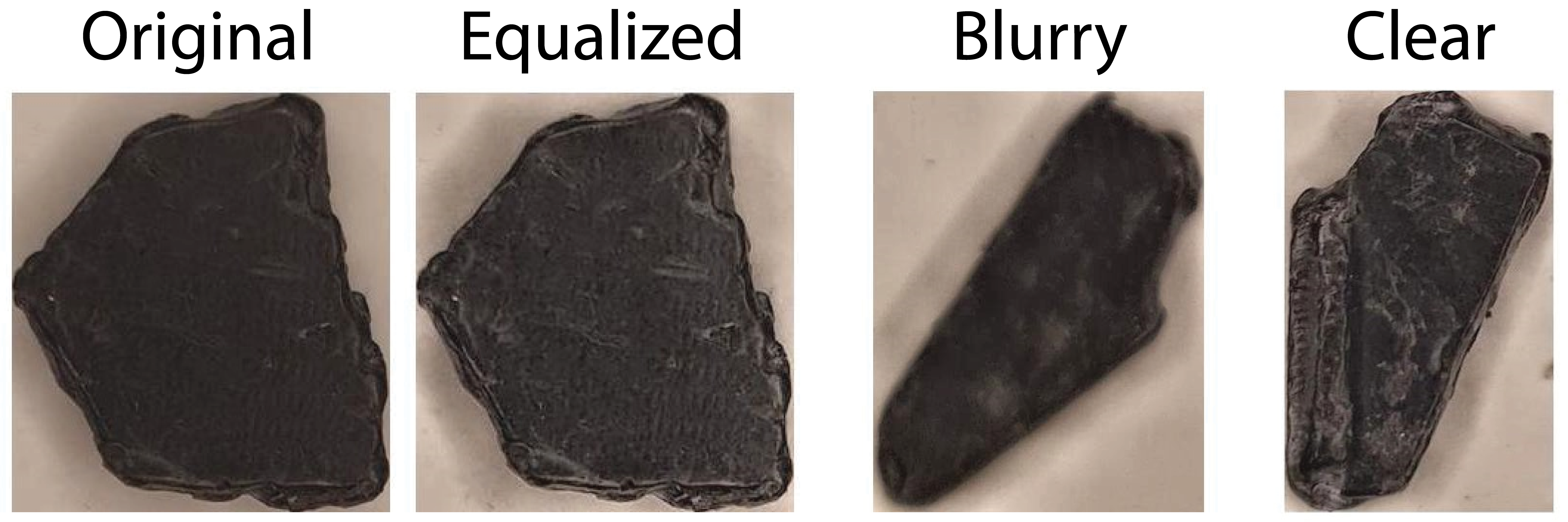}
        \caption{Different examples of segmented plastics.}
\label{fig11-12}
\end{figure}

\section{Methodology}

\subsection{Data Acquisition}

To avoid shadows caused by the plastics (which are already dark), we have designed an environment with sufficient lighting.
Another aim of this work is to evaluate the feasibility of a vision solution that employs a simple setup, built up with affordable hardware. 
Therefore, 
we use a ring light, the Avity Mobile Ring light, to obtain proper and uniform illumination. 
A ring light like this costs a few dozens of dollars (including associated hardware and tripod), and it is available in regular high-street stores.
The light is oriented to illuminate a flat acquisition surface perpendicularly from above, situated 7 cm away from the ring. The plastic pieces to be photographed are placed on a white-colored tray. 
The setup can be seen in Figure~\ref{fig8-15-16}, left. 
%
%
For image acquisition, we use the rear camera of a consumer smartphone, a Samsung S10.E, which has a dual camera of 12 MP (f1.5/f2.4 Dual Pixel Auto-Focus) + 16 MP (f2.2 ultra-wide).
The camera is fastened in the ring light grip, making sure it will not move during the different pictures.
Using a smartphone it is sufficient for this initial study, and it is readily available from the project team without the need of extra investments.
The camera stand is duct-taped to the floor so that it stays in the same position.
Example images of the two types of plastics employed in this work are shown in Figure~\ref{fig8-15-16}, center and right.

Before experiments can take place, pieces of plastics of known types are needed.
For this work, we have captured pictures of 215 pieces of ABS plastic and 215 pieces of PS plastic. 
To faithfully know the type of plastic appearing in the images, we have used a Bruker-Alpha spectrometer \cite{9} that uses MWIR illumination to determine the type of plastic (Figure~\ref{fig3-6-7}). 
The plastic piece is placed in the machine, which computes the MWIR scattering at different wavelengths (shown by the red line). The software of the machine \cite{10} then compares the obtained spectrum to a database of reference fingerprints, returning the closest type (blue line). 
If a reference with sufficient similarity is found, the software then determine that there is a match, and outputs the material type (in the example, it is a plastic of ABS type).
%


\begin{table*}[htb]
\normalsize
\begin{center}
\caption{Training results using different solvers.}
\label{tab2}
\begin{tabular}{|c|c|c|c|}
\hline
\textbf{Epochs=10, MiniBS=50} & \textbf{Validation Acc (Adam)} & \textbf{Validation Acc (sgdm)} & \textbf{Validation Acc (rmsprop)} \\ \hline \hline
\textbf{Test 1} & \textbf{88.33\%} & 78.3\% & 61.67\% \\ \hline
\textbf{Test 2} & \textbf{75\%} & 70\% & 65\% \\ \hline
\textbf{Test 3} & \textbf{85\%} & \textbf{85\%} & 80\% \\ \hline \hline
\textbf{Average} & \textbf{88.28\%} & 77.77\% & 68.69\% \\ \hline
\end{tabular}
\end{center}
\end{table*}


\begin{table*}[htb]
\normalsize
\centering
\caption{Training results using different data shuffle methods (solver: Adam).}
\label{tab3}
\begin{tabular}{|c|c|c|c|}
\hline
\textbf{Epochs=10, MiniBS=50} & \textbf{Validation Acc (Never)} & \textbf{Validation Acc (Once)} & \textbf{Validation Acc (Every epoch)} \\ \hline \hline
\textbf{Test 1} & 76.67\% & \textbf{81.62\%} & 81.55\% \\ \hline
\textbf{Test 2} & 73.33\% & 80\% & \textbf{86.67\%} \\ \hline
\textbf{Test 3} & \textbf{86.67\%} & 80\% & 83.33\% \\ \hline \hline
\textbf{Average} & 78.89\% & 80.54\% & \textbf{83.85\%}     \\ \hline
\end{tabular}
\end{table*}


\begin{table*}[htb]
\normalsize
\centering
\caption{Training results using different batch size and number of epochs (solver: Adam).}
\label{tab1}
\begin{tabular}{|c|c|c|c|c|c|c|}
\cline{1-3} \cline{5-7}
\textbf{epochs} & \textbf{MiniBS} & \textbf{Validation Accuracy} & \textbf{} & \textbf{MiniBS} & \textbf{epochs} & \textbf{Validation Accuracy} \\ \cline{1-3} \cline{5-7} 
40 & 20 & 83.54\%  & & 50 & 10 & 81.55\%  \\ \cline{1-3} \cline{5-7} 
40 & 30 & 88\%     & & 50 & 20 & 88.35\%  \\ \cline{1-3} \cline{5-7} 
40 & 40 & 90.29\%  & & 50 & 30 & 87.38\%  \\ \cline{1-3} \cline{5-7} 
40 & 50 & \textbf{94.17\%} & & 50 & 40 & \textbf{91.26\%}             \\ \cline{1-3} \cline{5-7} 
40 & 60 & 88.35\%  & & 50 & 50 & 86.56\%  \\ \cline{1-3} \cline{5-7} 
40 & 70 & 89.32\%  & & 50 & 60 & 85.24\%  \\ \cline{1-3} \cline{5-7} 
\end{tabular}
\end{table*}

\subsection{System Overview}

To segment the plastics, we first binarize the image using the the Otsu method \cite{[Otsu79]}. The binarized image is then plugged into Matlab´s \textit{bwlabel} function which label connected components of a binary image. Finally, the bounding box of each plastic found in the image is obtained with \textit{regionprops}.
A problem in segmenting objects which are dark would be the shadows, which would be considered as part of the object.
It could be circumvented by using shadow detection methods, e.g. \cite{5685658}. However, shadows will not be a problem in our work since our experiments are conducted in a strictly controlled environment, with the illumination source pointing perpendicularly from above (Figure~\ref{fig8-15-16}, left). 
After the objects (plastics) are located, they are saved as individual images.
An example of the segmentation process is shown in Figure~\ref{fig4}.

Convolutional Neural Networks (CNN) have over the last years become great at large-scale image recognition tasks, enabled by large public image databases such as ImageNet \cite{[Russakovsky15_ImagenetChallege]}. Here, we employ transfer learning by selecting a pre-trained architecture, and fine-tuning it to the type of images used in our application \cite{[Pan10]}. In many real-world applications, it is expensive or impossible to collect a sufficient amount of data to train the models from scratch. Transfer learning is thus a way to create new models with very little data compared to the initial training. The network architecture employed and open source scripts for retraining are provided by Matlab.


The network employed is AlexNet \cite{22}, initialized with the weights from the ImageNet challenge. This network obtained the 1$^{st}$ position in the ImageNet Large Scale Visual Recognition Challenge (ILSVRC) of 2012 \cite{[Russakovsky15_ImagenetChallege]}, being the first CNN-based solution to win the ILSVRC contest.
The network achieved a breakthrough in this competition, with a top-5 error
of 15.3\%, more than 10.8\% percentage points ahead of the runner up.
AlexNet has eight separate layers. The first five are convolutional layers, with some of them being followed by max-pooling layers, and ReLU as activation. Convolutional layers use filters on the image to create maps that contain features of the object in the image. In the end, three fully connected layers are used to classify the object in the image. The original network is trained in the context of the ImageNet challenge to classify 1000 classes of generic objects. 
Despite newer networks have gained in sophistication and depth, e.g. \cite{[He16],[Huang17]}, we believe that a simple network such as AlexNet will be sufficient for our purposes. Since we are using a reduced dataset and we want to classify only two classes, a more complex network would be more prone to over-fitting.

\section{Experiments and Results}

We employ the AlexNet pre-trained model in Matlab, with the last layer modified to classify only two classes.
Images of the individual plastics are segmented as indicated in the previous section and used as input of the CNN.
Since the images are non-square, they are enlarged horizontally or vertically to achieve a square image, and then, they are resized to the input size of the CNN (227$\times$227 pixels). 
To further enhance the contrast of the different images and compensate variability in local illumination, we apply Contrast-Limited Adaptive Histogram Equalization (CLAHE) \cite{[Zuiderveld94clahe]} using Matlab's \textit{adapthisteq} command.
The first two columns of Figure~\ref{fig11-12} show an example image of a plastic before/after applying CLAHE. 
200 pictures of each plastic type are used for CNN training (20\% of the pictures are used for validation), while the remaining 15 pictures are  set aside for testing once the optimal setup of the CNN is found via validation error analysis. 
The images should have as little noise as possible, meaning to be clear and detailed. 
The camera used sometimes loses focus due to the auto-focus function, resulting in blurry images (Figure~\ref{fig11-12}, third column). To solve this problem, the dataset has been manually reviewed before training, and such blurry images are discarded for training. 
Figure~\ref{fig11-12}, fourth column shows an example of a well-focused image.

The most significant settings of the CNN to fix are the solver, the mini-batch size, the maximum number of epochs, and the shuffle of data from epoch to epoch.
The mini-batch size specifies how many images are used per iteration of the training process. An epoch is when the whole training set has presented to the training algorithm. When it comes to the shuffle setting, it can be set to shuffle the training data at the beginning of every epoch, to shuffle it once (at the beginning of the first epoch), or to never shuffle it. Regarding the solvers, we have tested Adam, Stochastic Gradient Descent with Momentum (SGDM), and RMSProp. All their parameters (apart from those mentioned above) are set to the default parameters in Matlab. 

A series of test are conducted to find the best setting of the mentioned parameters. As metric, we use the validation accuracy at the end of the training, computed as the average of three runs of the training process.
Table~\ref{tab2} shows the results of the different solvers, with the mini-batch size set to 50 images, the maximum number of epochs to 10, and shuffling data on every epoch. As it can be seen, the Adam solver gives the best results consistently over the different training runs, so it will be set as the solver of choice for the remaining experiments of the paper.
In the next experiment, we test the different shuffle options (Table~\ref{tab3}) using the same mini-batch size and maximum number of epochs than the previous experiment. As expected, shuffling the data results in a better accuracy (preferably if it is shuffled on every epoch). 
Finally, in Table~\ref{tab1} we test different values of mini-batch size and maximum number of epochs, shuffling data on every epoch. Given a fixed number of 40 epochs (left part of the table), we observe that increasing the mini-batch size has a positive effect on the accuracy until a size of 50 is reached, after which the accuracy decreases. A similar effect is observed if we fix the mini-batch size and vary the maximum number of epochs.

\begin{figure}[htb]
\centering
        \includegraphics[width=0.45\textwidth]{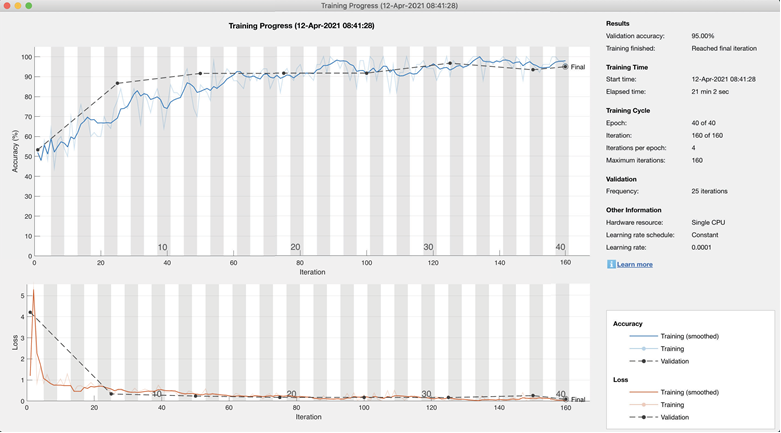}
        \caption{Training results with the best settings found (solver: Adam, mini-batch size: 50, maximum number of epochs: 40, data shuffle on every epoch).}
\label{fig13}
\end{figure}

From Table~\ref{tab1}, we can conclude that the optimal settings are: Adam solver, mini-batch size = 50, maximum number of epochs = 40, and shuffle data on every epoch. 
A further training run with this setup (Figure~\ref{fig13}) shows a validation accuracy of 95\% (consistent with Table~\ref{tab1} for this setup, which is of 94.17\% and 91.26\%).
Classification experiments are further conducted using the 15 pictures of each class type set aside for testing. The 15 images of ABS type are classified correctly (100\% accuracy), while 11 images of the PS type are classified correctly (73.3\%). Overall, 24 of 30 images are classified correctly, which makes an accuracy of 86.6\%.

\section{Conclusion}

Plastic waste from Electrical and Electronic Equipment (WEEE) is increasing rapidly, being one of the fastest growing waste streams. Given the hazardous nature of some materials found in electronic equipment, legislators are passing regulations that enforce a reduction in WEEE plastics that end up landfilled \cite{Buekens14}.
This paper has presented a system that makes use of a vision-based solution that employs Convolutional Neural Networks (CNNs) to automatize the classification of two types of black plastics, Polystyrene (PS) and Acrylonitrile Butadiene Styrene(ABS). 
These two types constitute the majority of plastics from electronics that are handled by the collaborating partner of this research, the STENA Nordic Recycling Center (SNRC) in Halmstad.
The AlexNet CNN architecture \cite{22} has been evaluated as classifier of these two types of plastics.
A dataset of 215 images of each plastic type has been also captured (Figure~\ref{fig8-15-16}). To do so, we have employed a ring light, oriented perpendicularly to a flat surface with a white background. Acquisition is done with the rear camera of a regular consumer smartphone. 
The proposed solution performs well in the described environment (indoor, controlled light and no motion between camera and plastics), with a validation accuracy of 95\%. Results with a separate test of 15+15 images gives an accuracy of 86.6\%.

The small amount of data employed may be an explanation of the different accuracy between validation and testing. A further look at the test results show that one of the plastics (PS type) is accumulating all the classification errors, while images of the other type are perfectly classified. 
An immediate avenue for improvement is thus to gather a bigger dataset to allow a better accuracy when training the network.
A camera with manual settings or with a better optics than a smartphone would also provide more control over the acquisition. 
In addition, we have captured the images in room with natural light entering through the windows, so it changed during the day. Despite the use of a light ring, the impact of other surrounding natural or artificial light would need a proper systematic study. 
A room without windows or the use of a black box that eliminates the impact of surrounding light would likely allow to attain better results. 
This is feasible in an industrial setting, where the plastics could be placed on a conveyor belt that passes through a light-controlled area.

Even if our initial results are promising, a bigger dataset would allow to test other more sophisticated networks that may provide a better accuracy. 
AlexNet is a serial network, but subsequent networks which surpassed its accuracy in ImageNet by a large margin include the popular ResNet \cite{[He16]} or DenseNet \cite{[Huang17]} variants. They have proposed concepts such as residual connections \cite{[He16]} and densely connected architectures \cite{[Huang17]}, which have allowed the training of deeper networks. 
These techniques have been improved further in networks such as MobileNetv2 \cite{[Sandler18mobilenetv2]}, whose aim is to reduce the size and computational budget of the network (making it less prone to over-fitting with reduced data) while exploiting recent developments in CNN architectures.
Another obvious improvement is to allow motion between the camera and the objects to be classified. This is because our solution is planned to be incorporated into a functional sorting facility where plastics are to end up into separate bags automatically. To do so, further research in detection and segmentation of moving plastics is necessary \cite{4} prior to presentation of images to the classifier.

\section*{Acknowledgment}

Author F. A.-F. thanks the Swedish Research Council (VR) and the Swedish Innovation Agency (VINNOVA) for funding his research. Authors also acknowledge the support of STENA Nordic Recycling Center (SNRC) in Halmstad.


\bibliographystyle{IEEEtran}

\bibliography{bibliography}

\end{document}